\pdfoutput=1
\documentclass[sigconf]{acmart}

\copyrightyear{2022}
\acmYear{2022}
\setcopyright{acmlicensed}\acmConference[KDD '22]{Proceedings of the 28th
ACM SIGKDD Conference on Knowledge Discovery and Data Mining}{August
14--18, 2022}{Washington, DC, USA}
\acmBooktitle{Proceedings of the 28th ACM SIGKDD Conference on Knowledge
Discovery and Data Mining (KDD '22), August 14--18, 2022, Washington, DC,
USA}
\acmPrice{15.00}
\acmDOI{10.1145/3534678.3539388}
\acmISBN{978-1-4503-9385-0/22/08}

\acmSubmissionID{rtfp1753}

\usepackage{subfigure} 
\usepackage{graphicx}  
\usepackage{float}  
\usepackage{bm}
\usepackage{multirow}
\usepackage{caption} 
\usepackage{xspace}
\usepackage{enumitem}
\usepackage{appendix}
\usepackage{hyperref}
\usepackage{xcolor}
\usepackage{nccmath}
\usepackage{balance}

\newcommand{\mname}{\texttt{M$^3$Care}\xspace}

\usepackage{tablefootnote}

\usepackage[linesnumbered,ruled,vlined]{algorithm2e}

\usepackage{etoolbox}
\SetKwInput{KwInput}{Input}                
\SetKwInput{KwOutput}{Output}              
\SetKwInput{KwTraining}{Training} 
\SetKwFor{KwFor}{For} 

\begin{document}


\title{\mname: Learning with Missing Modalities in Multimodal Healthcare Data}





\author{Chaohe Zhang}
\authornote{Both authors contributed equally to this research.}
\email{choc@pku.edu.cn}
\affiliation{%
  \institution{Key Lab of High Confidence Software Technologies, Ministry of Education}
  \city{}
  \country{}
}
\affiliation{%
  \institution{School of Computer Science, \\ Peking University}
  \city{Beijing}
  \country{China}
}

\author{Xu Chu}
\authornotemark[1]
\affiliation{%
  \institution{Department of Computer Science and Technology, Tsinghua University}
 \city{}
  \country{}
}
\affiliation{%
  \institution{Key Lab of High Confidence Software Technologies, Ministry of Education}
   \city{Beijing}
  \country{China}
}

\author{Liantao Ma}
\author{Yinghao Zhu}
\affiliation{%
  \institution{Key Lab of High Confidence Software Technologies, Ministry of Education}
  \city{}
  \country{}
}
\affiliation{%
  \institution{National Engineering Research Center of Software Engineering, Peking University}
  \city{Beijing}
  \country{China}
}


\author{Yasha Wang}
\authornote{Corresponding Author.}
\email{wangyasha@pku.edu.cn}
\affiliation{%
  \institution{Key Lab of High Confidence Software Technologies, Ministry of Education}
  \city{}
  \country{}
}
\affiliation{%
  \institution{National Engineering Research Center of Software Engineering, Peking University}
  \city{Beijing}
  \country{China}
}

\author{Jiangtao Wang}
\affiliation{%
  \institution{Center for Intelligent Healthcare, Coventry University}
  \city{Coventry}
  \country{UK}
}

\author{Junfeng Zhao}
\affiliation{%
  \institution{Key Lab of High Confidence Software Technologies, Ministry of Education}
  \city{}
  \country{}
}
\affiliation{%
  \institution{School of Computer Science,\\ Peking University}
  \city{Beijing}
  \country{China}
}

\renewcommand{\shortauthors}{Chaohe Zhang et al.} 
\begin{abstract}
  Multimodal electronic health record (EHR) data are widely used in clinical applications.
Conventional methods usually assume that each sample (patient) is associated with the unified observed modalities, and all modalities are available for each sample.
However, missing modality caused by various clinical and social reasons is a common issue in real-world clinical scenarios.
Existing methods mostly rely on solving a generative model that learns a mapping from the latent space to the original input space, which is an unstable ill-posed inverse problem.
To relieve the underdetermined system, we propose a model solving a direct problem, dubbed learning with \underline{M}issing \underline{M}odalities in \underline{M}ultimodal health\underline{care} data (\mname). 
\mname is an end-to-end model compensating the missing information of the patients with missing modalities to perform clinical analysis. 
Instead of generating raw missing data, \mname imputes the task-related information of the missing modalities in the latent space by the auxiliary information from each patient's similar neighbors, measured by a task-guided modality-adaptive similarity metric, and thence conducts the clinical tasks. 
The task-guided modality-adaptive similarity metric utilizes the uncensored modalities of the patient and the other patients who also have the same uncensored modalities to find similar patients. 
Experiments on real-world datasets show that \mname outperforms the state-of-the-art baselines. 
Moreover, the findings discovered by \mname are consistent with experts and medical knowledge, demonstrating the capability and the potential of providing useful insights and explanations.~\footnote{published as a conference paper in ACM SIGKDD 2022 (modified a few mistakes)}

\end{abstract}



\maketitle

\section{Introduction}
\label{sec:intro}

Multimodal data can provide complementary information from various modalities that reveal the fundamental characteristics of real-world subjects~\cite{xu2013survey,xu2018raim,hoang2021aid,baltruvsaitis2018multimodal,cai2018deep}.
Thus, many clinical applications, 
such as disease diagnosis and mortality prediction~\cite{ma2020concare,xu2018raim,ni2019modeling,zhang2018integrative,hoang2021aid}, 
require multimodal electronic health record (EHR) data to achieve good diagnostic or prognostic results. 
Conventional approaches usually assume that each sample is associated with the unified uncensored modalities, and all modalities are available for each sample~\cite{ma2021smil,zhang2019cpm}.
However, missing modality is a common issue in real-world clinical scenarios~\cite{huang2020fusion}. 
For example, different types of examinations are usually conducted for different patients~\cite{zhang2019cpm}.
Also, patients may lack some specific modalities due to patient dropout~\cite{pan2021disease}, sensor damage, data corruption~\cite{chen2020hgmf}, safety considerations~\cite{zhou2020hi,ramos2004mri} and high cost~\cite{ford2000non}.
Formally, we define modality-missing in EHR data as, for a sample, data for at least one modality is missing. 
The absence of a modality means that all features in this modality are missing.
Moreover, the modality-missing patterns (i.e., combinations of available modalities) make it more complex for the data with more modalities~\cite{zhang2019cpm}.

\begin{figure}[]
  \centering
  \includegraphics[width=\columnwidth]{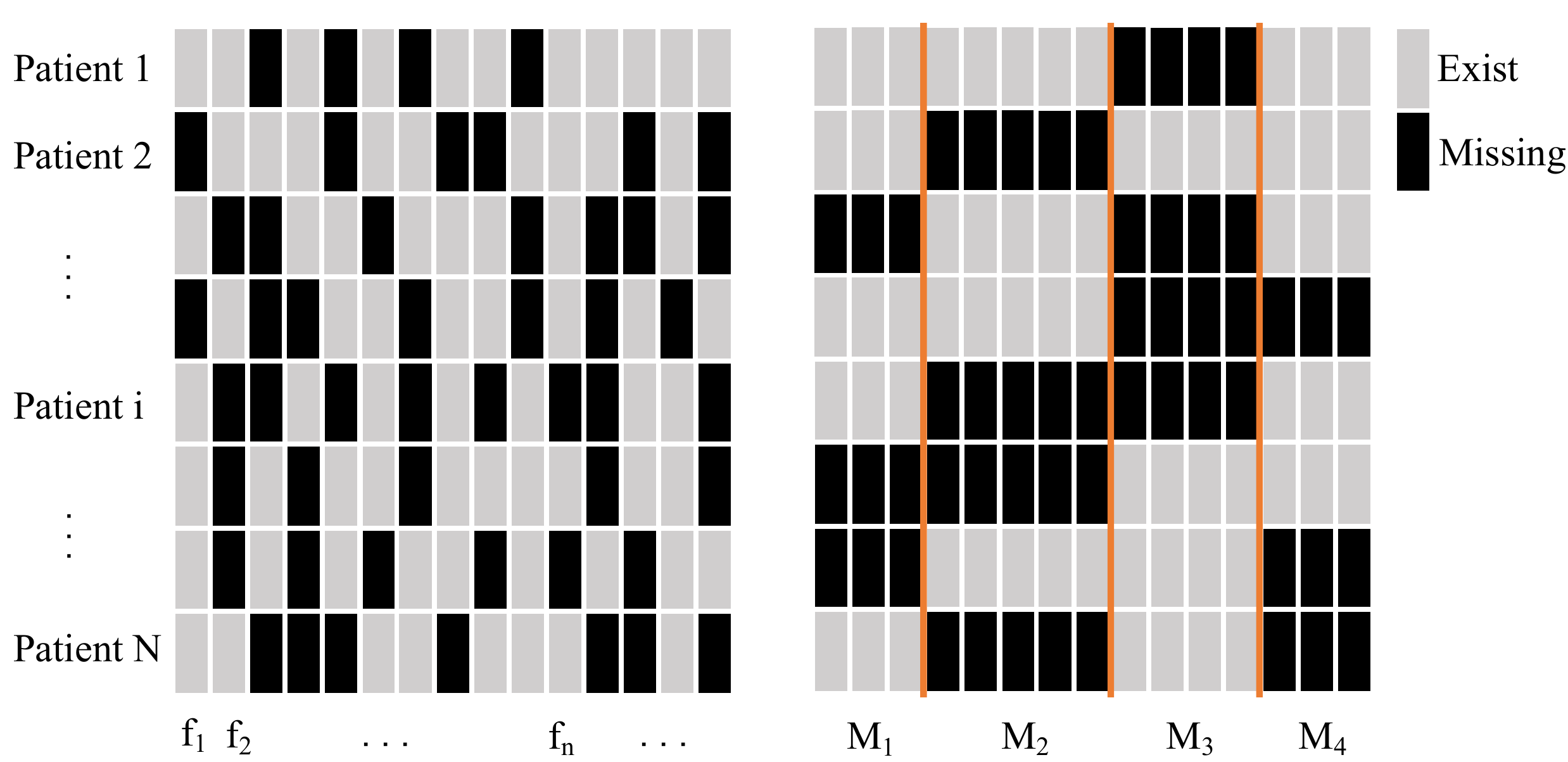}

  \caption{Left: missing features; right: missing modalities. Each row refers to a patient. In the left figure, each column refers to a feature ($f_n$). In the right figure, each group of columns refers to a modality ($M_n$), meaning that a modality contains many features. These features have high correlations, since they belong to the same modality (e.g., medical images, medical notes, etc.). The boxes in gray indicate the features exist, and others in black represent missing ones. 
  }
  \label{fig:med_missing_modal}

\end{figure}

Thus, some pioneering research works are proposed to handle the modality-missing issue.
Some researchers drop the incomplete samples~\cite{ni2019modeling,wang2020multimodal} and achieve some improvements.
However, this approach cannot be applied in areas where data is scarce and contains rigid requirements, such as healthcare.
Also, it will escalate the small-sample-size issue and over-fitting~\cite{chen2020hgmf,pan2021disease}.

The complementary way to dropping methods is the imputation-based method.
As shown in the left part of Figure~\ref{fig:med_missing_modal}, some methods assume that the entries of the data matrix are missing at random (or some more specific assumption on the matrix space, e.g., incoherence, confer~\cite{davis2006relationship} for a survey), and the missingness can be imputed via modeling correlations between the columns (features)~\cite{lin2020missing}.
Whereas, as illustrated in the right part of Figure~\ref{fig:med_missing_modal}, missing modalities manifest themselves by column-wise consecutive missingness, where the most correlated information inside the same modalities is missing entirely. On the other hand, the features inside the same modality are naturally more correlated than thereof in different modalities, exhibiting a coherent behavior in the matrix space. Thus the traditional imputation methods do not work well~\cite{wang2020multimodal}.
Additionally, block-wise missingness~\cite{yi2016st}, such as image~\cite{hong2019deep} or geosensory data~\cite{yi2016st}, often assumes a sample realization is from the matrix space, such that the row vectors in a matrix are not permutable.
Different from that, the rows in missing modalities refer to the samples, which are permutable. 
This results in the prior spatial (or spatio-temporal) correlations required to complete the block-wise missingness are not present in the missing modalities.
In a word, conceptually, there is a gap between the missing modalities (column-wise consecutive missingness) and the existing imputation-based methods for random or block-wise missingness.

In methodology, some deep generative methods~\cite{ngiam2011multimodal,li2020estimation,pan2021disease,cai2018deep} are proposed for missing modalities.
Essentially, such methods are usually based on the manifold assumption: the probability mass of real-world objects is supported on low-dimensional manifolds~\cite{agarwal2010learning}. 
In terms of EHR data imputation, the manifold assumption can be interpreted as the low-rankness and stability of the feature covariance matrix. 
In other words, there exist a set of low-dimensional basis vectors and a deterministic mapping $\mathcal{T}$ subject to: 
(a) the basis vectors span the pre-image of the mapping $\mathcal{T}$, 
(b) and the observed EHR feature vectors live on the image of the mapping $\mathcal{T}$. 
Existing EHR generative-based completion methods tackle the problem by solving a generative model which learns a mapping from a latent space (spanned by basis vectors) to the original input space~\cite{ngiam2011multimodal,li2020estimation,pan2021disease,cai2018deep}. 
Solving such mapping is essentially an ill-posed inverse problem (low- to high-dimensional underdetermined system~\cite{datta2010numerical}), whose solution is often non-unique and unstable~\cite{kabanikhin2008definitions}. 
On the other hand, 
such complex auxiliary models may introduce extra noise, which has negative impacts~\cite{chen2020hgmf,wang2020multimodal,enders2010applied}.
To this end, the problem of completing missing modalities requires a different way.

\begin{figure}[]
  \centering
  \includegraphics[width=\columnwidth]{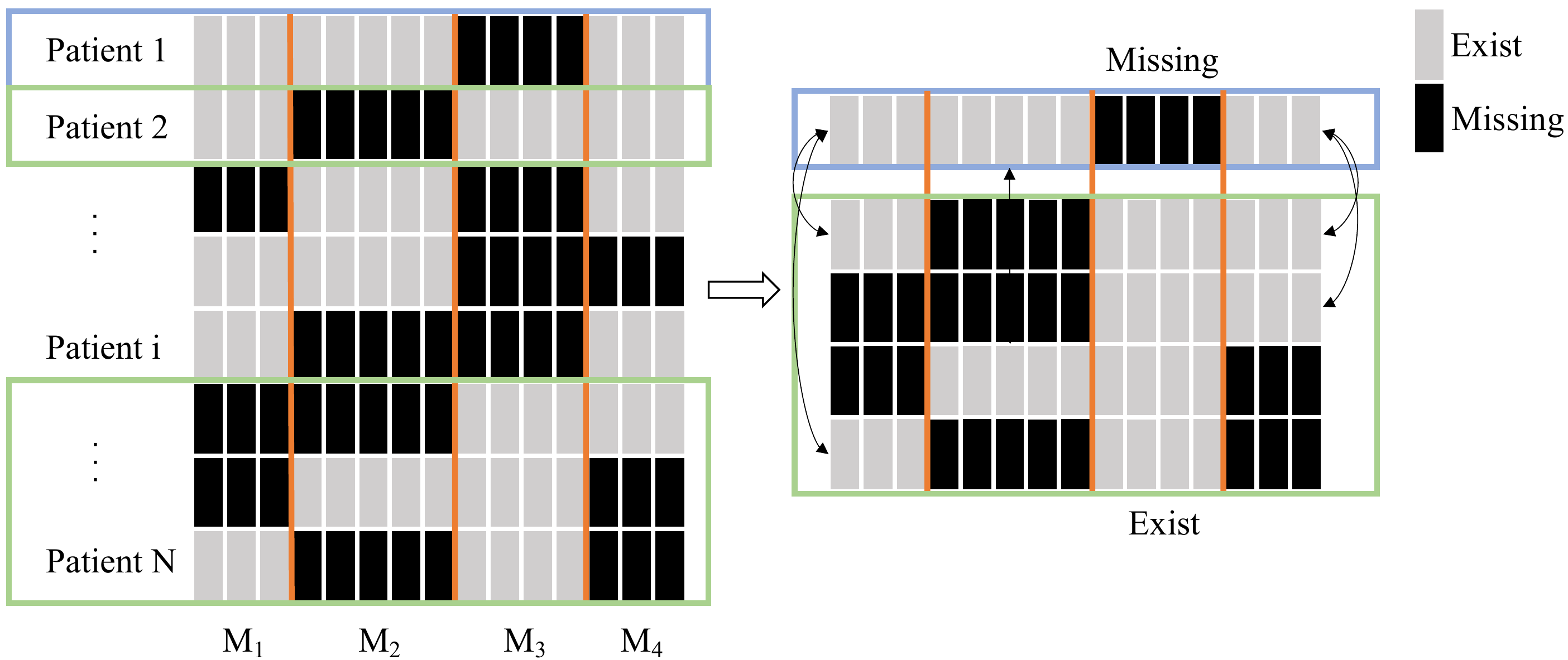}

  \caption{Intuition: For a patient with missing modalities, we utilize the other uncensored modalities of the patient and the other patients who also have the same uncensored modalities to find similar patients and estimate the missing information. }
  \label{fig:insight}

\end{figure}

In fact, for missing modalities, solving the generative model is not necessary. 
By assuming the low-rankness and the stability of the covariance matrix of EHR features, the locally similar row vectors (patients with similar features from some uncensored modalities) in a sub-matrix imply globally similar row vectors (those patients should have similar missing features) in the data matrix.
Moreover, if a local row vector $X$ falls in a convex hull spanned by a set of local row vectors, then the global row vector corresponding to $X$ is likely to fall into the convex hull spanned by the particular set of global row vectors.
Thus, instead of solving the inverse problem of the low- to high-dimensional mapping, we can solve a less underdetermined problem: comparing similarities of local row vectors and impute the missing entries in a row by referring to the uncensored entries of locally similar rows. 
The similarity comparison can be conducted in the original input space with a sophisticated metric on data manifolds, or in a learned latent space with a more straightforward metric.
More importantly, modeling the similarity relationship in the low-dimensional latent space is a direct (or forward, namely) problem of solving a mapping from a high-dimensional space to a low-dimensional space, which is less complicated than the inverse problem of solving a generative model.

On the other hand, this intuition is also in agreement with the real-world clinical practice, i.e., how doctors use the relationships between patients to assist the clinical analysis\footnote{\hyperref[sec:discovery]{We also substantiate this intuition by mining the real-world clinical datasets, please refer to the Intuition discovery experiments in Appendix.}}.
If two patients are similar in one clinical modality, they are more likely to be similar in another one~\cite{bari2009randomized,kwiecinski2021native,gong2018learning,huang2018cross}.
Thus, as shown in Figure~\ref{fig:insight}, for a patient with missing modalities, we can utilize the other complete modalities of the patient (local row vector) and other patients who also have these modalities (local uncensored row vectors), to find similar patients. 
Although seeming straightforward, applying this intuition to clinical tasks will face the following challenges:

\textbf{Challenge 1.
Which space can be used to perform imputation?}
There are at least two options: the original input space or a learned latent space.
Existing methods usually estimate missing data in the original input space~\cite{ngiam2011multimodal,cai2018deep,li2020estimation}.
However, the probability distribution in the original input space contains task-relevant and task-irrelevant information. Imputation in the original input space treats task-irrelevant information equally, weakening the task-specific information conveyed in EHR data.
This results in an indiscriminate loss of task-relevant and task-irrelevant information, leading to inferior performance.

\textbf{Challenge 2.
What metric(s) should be used to model the similarity relation?}
T-LSTM~\cite{baytas2017patient} uses autoencoders to generate patient representations, based on which the similarity is obtained. 
SMIL~\cite{ma2021smil} uses multivariate Gaussian to assign similarity weights.
However, they did not associate the connotation of similarity with clinical tasks.
In different clinical tasks (e.g., mortality prediction and disease diagnosis), the patient characteristics that need attention are different, so two patients considered similar in one clinical task may be considered not so similar in another~\cite{zhang2021grasp}.
More importantly, the similarity metric might vary for different modalities. 
The manifolds in different modalities in the original input spaces are naturally equipped with different metrics. 
A learned deterministic mapping from the original input spaces to the learned latent space is not guaranteed to result in a unified metric for different mapped modalities in the latent space.

\textbf{Challenge 3.
How to infer the local-to-global similarities of patients?}
Metrics computing the similarities between local row vectors is not sufficient to describe the similarities of global row vectors. 
Thence modeling the intra- and inter-correlations of features from various modalities is challenging but indispensable to aggregate the information from different modalities.

By jointly considering the above issues, we propose a model learning with \underline{M}issing \underline{M}odalities in \underline{M}ultimodal health\underline{care} data (\mname), an end-to-end inductive learning model to compensate for the missing modalities and perform clinical tasks.
In summary, our main contributions are summarized as follows:
\begin{itemize}[leftmargin=*,noitemsep,topsep=2pt]

    \item We propose \mname to compensate for 
    the modality-missing patient in the latent space and perform clinical tasks with EHR data.
    Since the latent representations are highly compressed and task-supervised, this results in less loss of task-relevant information and is thus more beneficial for subsequent tasks in an end-to-end learning schema (Response to Challenge 1). 
    
    \item Methodologically, (a) \mname uses task-guided deep kernels in the latent space of each modality as the metric to compute patient similarities (Response to Challenge 2). (b)
    \mname captures intra-correlations within each modality and inter-correlations between modalities by a self-attentive multi-modal interaction module so that the local metrics are aggregated to calculate the similarities of global row vectors. (Response to Challenge 3).
 
    \item Extensive experiments show that \mname outperforms all state-of-the-art models under multiple levels of incompleteness in different evaluation metrics. Besides, the findings discovered by \mname are in accord with experts and medical knowledge, which shows it can provide useful insights and explanations.
\end{itemize}

\section{Related Work}
\noindent\textbf{Multimodal learning for healthcare.} 
With the advancement of medical technology, comprehensive healthcare is burgeoning to meet the demands of patients.
This has allowed for multiple medical modalities (e.g., medical image, clinical notes, etc.) to be analyzed to offer patients with feedback, as well as physicians with insights on clinical applications~\cite{xu2018raim,ni2019modeling,hoang2021aid,gao2020compose,ma2020adacare}.

To this end, multimodal learning for healthcare has attracted the interest of researchers.
For example, RAIM~\cite{xu2018raim} is proposed for jointly analyzing continuous monitoring data (e.g., ECG, heart rate) and discrete clinical events (e.g., intervention, lab test) to predict patient decompensation.
~\citet{gao2020compose} utilize a multimodal inference model to jointly encode trial criteria text and patient EHR tabular data for patient-trial inference.
~\citet{huang2020multimodal} develop and compare different multimodal fusion architectures to classify Pulmonary Embolism (PE) cases.
\citet{ma2021distilling} and~\citet{hoang2021aid} develop distillation frameworks to leverage the multimodal EHR data to enhance the prognosis.
Although the methods above work well, one common drawback is that they can only handle samples with complete modalities.
Limitations exist while modeling multimodal interactions with the presence of missing modalities.

\noindent\textbf{Methods for missing modalities.}
Currently, there have been research interests in handling missing modalities, which are mainly divided into two types: deleting incomplete samples or imputing missing modalities.
For the first type, FitRec~\cite{ni2019modeling} performs workout profile forecasting based on multimodal user data, which discards samples with missing modalities.
~\citet{wang2020multimodal} propose a knowledge distillation framework on samples with complete modalities, while distilling the supplementary information from the incomplete ones.
However, such methods can not handle the samples in need with missing modalities and have limitations to be applied in rigid demand domains like healthcare.
Besides, such methods dramatically reduce training data and result in over-fitting of deep learning models~\cite{chen2020hgmf,pan2021disease},
especially when there are many modalities and many different missing combination patterns\footnote{e.g., five modalities can result in $2^5-1 = 31$ missing patterns.}.

The second type is generating the missing modalities at first~\cite{pan2021disease,ma2021smil,cai2018deep}.
However, the incompleteness of modalities leads to column-wise consecutive missingness of features, which makes traditional methods like matrix completion can not be used~\cite{wang2020multimodal}.
Some advanced generative methods such as autoencoders~\cite{ngiam2011multimodal,li2020estimation} and generative adversarial networks (GAN)~\cite{pan2021disease,cai2018deep} have been proposed. 
These solutions, however, may introduce unwanted extra noise~\cite{chen2020hgmf}.
Especially when the size of samples with complete modalities is small, yet the number of modalities is large, the modalities imputed by such methods may have a negative effect~\cite{wang2020multimodal,enders2010applied}.
Moreover, while facing data with many modalities or missing patterns, the number of generators required is also large, which is difficult to train.
~\citet{chen2020hgmf} propose a method to enable multimodal fusion of incomplete data and get good performance.
However, the method is transductive and needs pre-training, indicating difficulty in applying it when a new sample comes.
Therefore, in this paper, we propose a new inductive framework to tackle the above limitations in an end-to-end schema.

\section{Problem Formulation}


In this section, we define the input data and the modeling problem in this paper. 
Besides, the necessary notations used in the paper are listed in Table~\ref{tab:notations} for ease of understanding.

\noindent\textbf{Definition 1 (Patient multimodal EHR data).}
In multimodal EHR data, each patient can be represented as a collection of observations from multiple modalities (data sources), e.g., medical images, clinical notes, lab tests, etc. 
Suppose $M$ is the number of modalities, $N$ is the number of patients (samples), and let $n$ be the subscript referring to a specific patient, the patient multimodal EHR dataset can be denoted as: $\mathbf{\mathbb{X}} = \{{\mathbf{X}}_{n}\}_{n=1}^N = \{( \mathbf{x}_{n}^{{1}}, \mathbf{x}_{n}^{{2}}, ..., \mathbf{x}_{n}^{{M}} )\}_{n=1}^N$.

\noindent\textbf{Definition 2 (Patient data with missing modalities).}
For a specific patient, as mentioned in Section~\ref{sec:intro}, various clinical and social reasons cause the absence of some modalities.
Thus, the observed data of a patient are represented as: ${\mathbf{X}}_{n} = \{ \mathbf{x}_{n}^{{1}}, \mathbf{x}_{n}^{{2}}, ..., \mathbf{x}_{n}^{{M^{'}}} \}$, where $0 < M^{'} < M$.
It should be noted that, we used the most relaxed setting, i.e., the modality missingness is irregular across patients. 
In all the training and test (validation) sets, each modality is potentially missing, but at least one modality is present for each patient.

\noindent\textbf{Problem 1 (Disease diagnosis).}
Given a patient’s multimodal EHR data ${\mathbf{X}}_{n}$ with some missing modalities, we formulate the disease diagnosis task as a \textit{binary} or \textit{multi-label} classification problem, which is diagnosing the disease $\ \mathbf{y} \in \{0,1\}^{|C|}$ of the patient, where $|C|$ is the number of the unique number of disease categories.

Table~\ref{tab:notations} shows the notations used in the paper.
 \begin{table}[h]
    \centering
    
      \caption{Notations for \mname}
    \label{tab:notations}
    \resizebox{\columnwidth}{!}{
    \begin{tabular}{c|l}
        \hline
        Notation & Definition \\
         \hline
         $ \mathbf{y} \in \{0,1\}^{|C|} $ & Ground truth of the classification target\\
         $\hat{\mathbf{y}} \in [0,1]^{|C|}$ & Classification result\\
         $\mathbb{X}$ & The multimodal EHR dataset \\
         $\mathbf{{X}}_{n} \in \mathbb{X}$ & The $n$-th patient in the dataset \\
        $\mathbf{x}_{n}^{{m}}$ & Modality $m$'s raw data of the $n$-th patient \\

         \hline
         $\mathbf{h}^{m}_{n} \in \mathbb{R}^{N_{h}}$ & Learned representation of modality $m$ of patient $n$\\
         ${\mathbf{H}^{{m}}} \in \mathbb{R}^{B \times \bm{N}_{h}}$ & Learned representations of modality $m$ for the patient batch \\
         ${\hat{\mathbf{H}}^{m}} \in \mathbb{R}^{B \times \bm{N}_{h}}$ & Modality $m$'s auxiliary information representation  \\
         & aggregated from similar patients\\
         $\bar{\mathbf{H}}^{\text{seq}} \in \mathbb{R}^{N_{\text{seq}} \times \bm{N}_{h}}$  & Representations of a sequential modality with positional \\
         & encoding added\\

         \hline
         $\delta\in (0, 1)$ & Parameter to control learnable kernel \\
         $ \Pi^{{m} } \in\mathbb{R}^{B\times B}$ & The patient similarity matrix for modality $m$\\
         $\text{mask}^{m}  \in\mathbb{R}^{B\times B}$& Matrix of booleans that determines each element of the associated \\
         &  value is valid or not (i.e., similarity for missing modality is invalid) \\
         $\Lambda \in\mathbb{R}$ & A learnable threshold to filter out dissimilar pairs \\
         $ \tilde{\Pi} \in\mathbb{R}^{B\times B}$ & The comprehensive similarity matrix across all modalities\\

         $\mathbf{z}^{l} $ & The output representations of the $l$-th layer in the model \\
         $\tilde{\mathbf{z}}^{l} $ & The middle representations inside the $l$-th layer in the model \\
         $\alpha_{m} \in\mathbb{R}^{B \times 1}$ & The importance of self-information ${\mathbf{H}^{{m}}} $\\
         $\beta_{m} \in\mathbb{R}^{B \times 1} $ & The importance of similar patients information ${\hat{\mathbf{H}}^{m}} $\\

         \hline
    \end{tabular}}
\end{table}
\section{Methodology}
\subsection{Overview}
\begin{figure*}[t]
\centering
\includegraphics[scale=0.48]{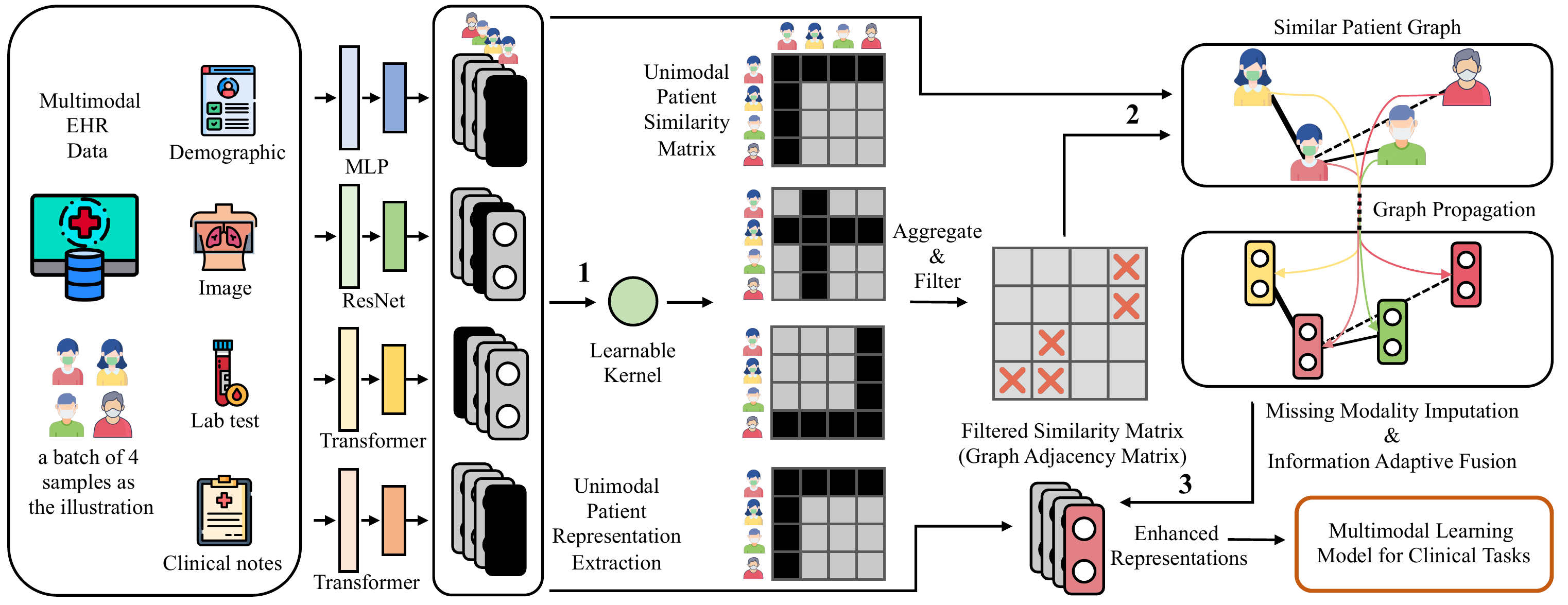}
\caption{Framework of \mname, the black boxes denote missing.
}
\label{fig:framework}
\end{figure*}

\begin{figure}[t]
\centering
\includegraphics[scale=0.5]{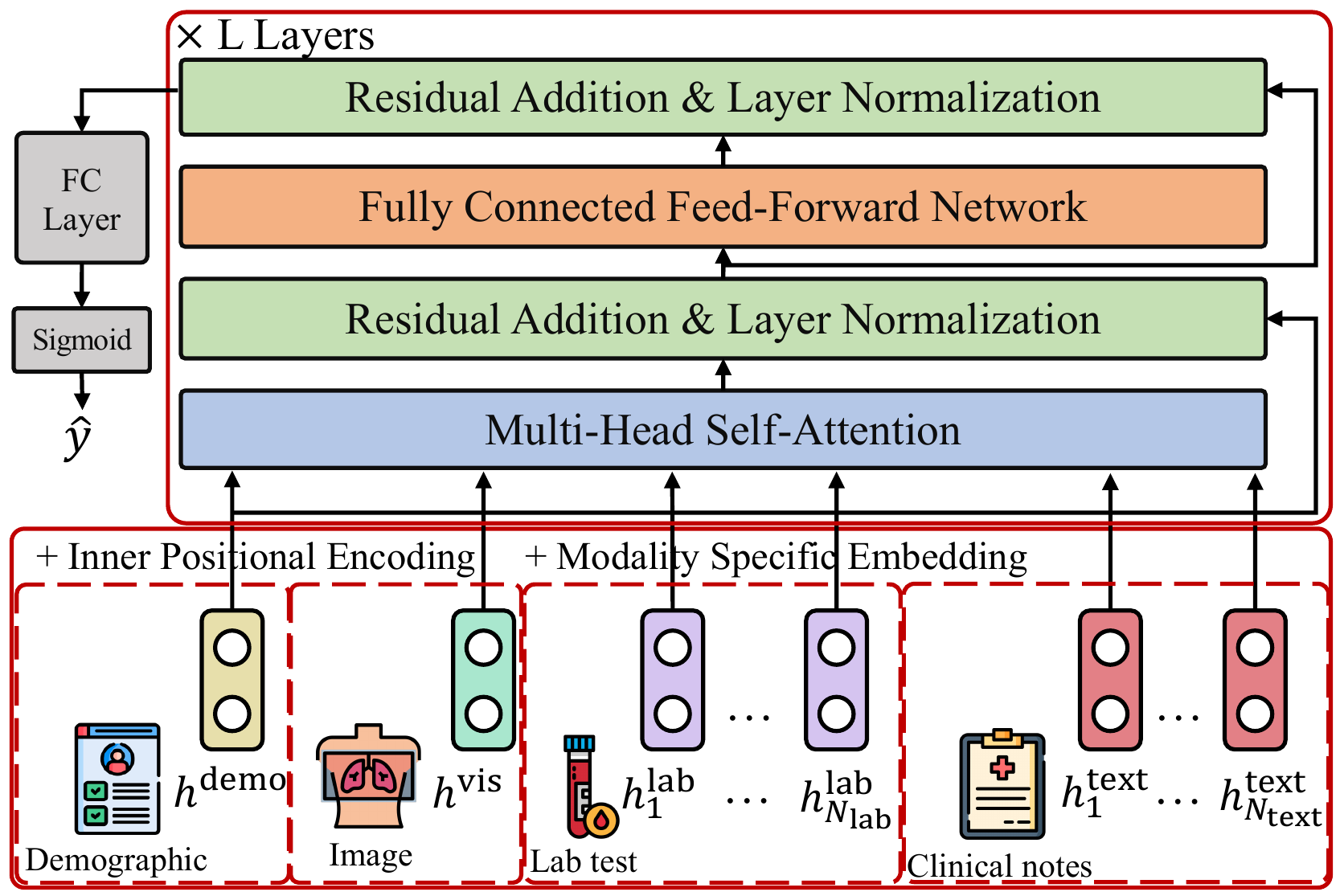}
\caption{Multimodal Learning Model for Clinical Tasks. Continued from the bottom right corner in Figure~\ref{fig:framework}. 
     }
\label{fig:framework2}
\end{figure}

Figure~\ref{fig:framework} and Figure~\ref{fig:framework2} show the architecture of \mname. 
It consists of two main sub-models.
The first one is used to compensate for the missing information in the latent space (Corresponding to Figure~\ref{fig:framework}).
The other utilizes the processed representations to perform the clinical tasks (Corresponding to Figure~\ref{fig:framework2}).
Specifically, \mname includes the following detailed components:

\begin{itemize}[leftmargin=*,noitemsep,topsep=2pt]
    \item The \textit{Unimodal Representation Extraction} module maps the original input features of a patient in each modality to the latent space by encoders with various backbones.
    The backbones are different due to different input modalities (Left part in Figure~\ref{fig:framework}).
    
    \item The \textit{Similar Patients Discovery and Information Aggregation} module computes patient similarities 
    of each modality with learned task-guided deep kernels. The similarities induce patient graphs. With graph information propagation, the information from similar patients is aggregated (Middle and right parts in Figure~\ref{fig:framework})

    \item The \textit{Adaptive Modality Imputation} module imputes the missing modality in the latent space with the aggregated information, and fuses the existing modality and the auxiliary information to enhance the representation learning (Right part in Figure~\ref{fig:framework}).
    
    \item The \textit{Multimodal Interaction Capture} module takes the intra- and inter-modality dynamics into consideration to perform the final clinical tasks  (Figure~\ref{fig:framework2}).
\end{itemize}

\subsection{Unimodal Representation Extraction}
For a specific patient $n$, it is hard to model the interactions among the raw data, since his/her data $\mathbf{X}_{n}$ is high-dimensional and inconsistent with respect to different data structures in different modalities~\cite{chen2020hgmf}.
Therefore the unimodal representation extraction models are in need to extract the task-relevant feature latent representation in the latent space of each modality.
Here, suppose $f_{m}\left(\cdot ; \boldsymbol{\Theta}_{m}\right)$ be the modality $m$'s unimodal representation extraction model with learnable parameter $\Theta_{m}$.
For modality $m$'s raw input, the corresponding latent representation can be obtained via:
\begin{equation}
\label{eq:hm}
\mathbf{h}^{m}_{n}=f_{m}\left(\mathbf{x}^{m}_{n} ; \boldsymbol{\Theta}_{m}\right),
\end{equation}
where $\mathbf{h}^{m}_{n} \in \mathbb{R}^{N_{h}}$ and $N_{h}$ is the dimension of the representation for modality $m$.
We use the lowercase letter $\mathbf{x}^{m}_{n}$ to denote a modality for a single patient $n$.

Three types of $f_{m}\left(\cdot ; \boldsymbol{\Theta}_{m}\right)$ are taken into consideration in this paper: 
1) ResNet~\cite{he2016deep} for embedding image modalities; 
2) Transformer Encoder~\cite{vaswani2017attention} for embedding sequential modalities, such as time-series data like patient laboratory test, medication, and free texts like clinical notes; 
and 
3) multi-layer perceptron (MLP) for embedding vector-based modalities, such as demographic information.
As shown in the left part in Figure~\ref{fig:framework}, the black boxes denote the missing information.

\subsection{Similar Patients Discovery and Information Aggregation}
The above section describes how to deal with a single sample, and now we focus on a group of samples.
Given a batch of patients, the representations of modality $m$ for them are denoted as:
\begin{equation}\nonumber
\label{eq:batchH}
\mathbf{H}^{m} = [ \mathbf{h}_{1}^{m}, \mathbf{h}_{2}^{m}, \,\dots,  \mathbf{h}_{B}^{m}]^{\intercal} \in\mathbb{R}^{B \times N_{h}},
\end{equation}

where $B$ is the batch size. 
For sequential modalities, we select the representation of the last timestamp as the one.

For a specific patient with missing modalities, as shown in Figure~\ref{fig:med_missing_modal}, the representations from the missing modalities cannot be obtained, which results in a lack of modality information.
As discussed in Section~\ref{sec:intro}, we can compare similarities of local patient data and impute the missing modalities of the patient by referring to the uncensored modalities of locally similar patients.
There is thus a problem with similar patient discovery: What metric(s) should be used to model the similarity relation?
In practical applications, one often adopted strategy is to try different kinds of similarity measures on the learned representations in each modality space, such as Cosine, Euclidean distance, and so on, and then select the best similarity measure~\cite{kang2016top}.
However, this approach is time-consuming, and even if one tries different similarity metrics, it is found that those traditional similarity metrics often fail to consider the local environment of data points, and may learn incomplete and inaccurate relationships.
In this case, it is unlikely that the traditional similarity function will be adequate to capture the local manifold structure precisely~\cite{kang2017kernel}. 
Moreover, complex relationships such as higher-order statistics are failed to capture in that way~\cite{kang2017kernel}.

To this end, we extend this idea to kernel spaces and select the RBF kernel.
Given two samples $\mathbf{h}_{i}^{m}$ and $\mathbf{h}_{j}^{m}$, the similarity calculated from the RBF kernel is defined as:
\begin{equation}
\label{eq:GKernel}
    k(\mathbf{h}_{i}^{m}, \mathbf{h}_{j}^{m})=\exp (-\frac{\|\mathbf{h}_{i}^{m}-\mathbf{h}_{j}^{m}\|_{2}^{2}}{2 \sigma^{2}}),
\end{equation}
where $\sigma$ is the bandwidth to control the extent to which similarity of small distances is emphasized over large distances. 
Following~\cite{chu2020distance}, we set $\sigma$ as a fraction of the mean distance between examples.
Expanding the exponential via Taylor series, we can see that the RBF kernel implies an infinite dimension mapping, capturing the higher-order statistics.

Furthermore, as mentioned in Challenge 2, since the data are in multiple modalities, \mname is required to calculate similarities in each modality space and associate the connotation of similarity with clinical tasks.
Thus, a task-guided modality-semantic-adaptive similarity metric is needed.
We extend the standard RBF kernel to deep kernel~\cite{liu2020learning} to build an adaptive kernel with a learnable network to fit the representations in a modality data-driven way.
Specifically, the kernel is denoted as:
\begin{equation}
\label{eq:}
    k_{\omega_{m}}(\mathbf{h}_{i}^{m}, \mathbf{h}_{j}^{m})=[(1-\delta_{m}) k(\phi_{\omega_{m}}(\mathbf{h}_{i}^{m}), \phi_{\omega_{m}}(\mathbf{h}_{j}^{m}))+\delta_{m}] q(\mathbf{h}_{i}^{m}, \mathbf{h}_{j}^{m}),
\end{equation}
where $\phi_{\omega_{m}}$ is a network with parameters $\omega_{m}$ for modality $m$. $k$ and $q$ are different RBF kernels with different $\sigma$.
The $\delta_{m} \in (0, 1)$ is a learnable \textit{safeguard} to preventing the learned kernel from going extremely far-away from the right direction.

Now, back to the batch of patient representations, the pairwise similarities with respect to each modality are calculated as (i.e., the No.1 arrow in Figure~\ref{fig:framework}):
\begin{equation}
\label{eq:pim}
\begin{split}
    \Pi^{m }& =k_{\omega_{m}}(\mathbf{H}^{m} , {\mathbf{H}^{m}}), \\
\end{split}
\end{equation}
where $\Pi^{m} \in\mathbb{R}^{B\times B}$ is the patient similarity matrix for modality $m$.
Meanwhile, to ensure the stability of the similarity measure and prevent collapse, we restrict the norm of the difference between the deep representation and the original representation as the optimization objective:
\begin{equation}
\label{eq:norm_loss}
    \mathcal{L}_{\text{stab}} = 
\sum_{m=1}^{M} \left|\|\phi_{\omega_{m}}(\mathbf{H}^{m})\|_F - \|\mathbf{H}^{m}\|_F \right|,
\end{equation}
where the outer $\left|\cdot \right|$ means the absolute value, and the inner $\|\cdot\|_F$ is the Frobenius norm.

Because of the characteristic of the kernel method, $\Pi^{m}$ is a totally positive symmetric matrix, and each cell $\Pi^{m}_{i,j}$ in $\Pi^{m}$ ranges from 0 to 1, denoting the similarity between the $i$-th patient and $j$-th patient for modality $m$.
However, in this way, all the patients are considered similar since the positive similarity. 
There should be dissimilar ones to be filtered out.
A straightforward way is setting a threshold, and the similarities below the threshold are considered dissimilar.
Nevertheless, in the early training phase, we notice that when the model is not convergent and the representations are not fully learned, all similarities are unstable.
The threshold may filter out some similar patients and lead to inferior performance.
Moreover, the determination of the value of the threshold is not trivial.
Thus, we utilize a more flexible learnable threshold here.
By comprehensively considering similarity from each modality, the filtered similarity matrix can be obtained as:
\begin{equation}
\label{eq:pi}
\begin{split}
   \tilde{\Pi} =  \frac{\sum_{1 }^{M} \Pi^{m} \cdot  \text{mask}^{m}  }{\sum_{1 }^{M} \text{mask}^{m} + \epsilon  } 
\end{split}
\end{equation}
\begin{equation}
\label{eq:piij}
\begin{split}
   \tilde{\Pi}_{i,j}  = \begin{cases}
  & \tilde{\Pi}_{i,j}   \ \  \text{ if } \tilde{\Pi}_{i,j} >  \Lambda   \\
  & 0  \ \ \ \ \ \ \text{ if  } \tilde{\Pi}_{i,j} \le \Lambda  
\end{cases}
\end{split}
\end{equation}
where $\Lambda $ is the learnable threshold to filter out dissimilar pairs and $\epsilon$ is used to prevent unstable division by zero.
$\text{mask}^{m}  \in\mathbb{R}^{B\times B}$ is the mask matrix of booleans that determines whether each element of the associated value is valid or not (i.e., similarity for missing modality is invalid).
For example, in modality $m$, if both the data of the $i$-th sample and $j$-th sample exist, $\text{mask}^{m}_{i,j} = 1$, and otherwise $\text{mask}^{m}_{i,j} = 0$, which masks the invalid similarity cell.

Our aim is to impute the modality-missing sample by incorporating auxiliary information from the similar patients.
Thus, to aggregate the information from the similar ones, we formulate the batch of patients' representations as a graph in each modality, with the similarity matrix $\tilde{\Pi}$ as the graph adjacency matrix (i.e., the No.2 arrow in Figure~\ref{fig:framework}).
Then the graph convolutional layers (GCN)~\cite{kipf2016semi} are applied to enhance the representation learning by leveraging the structural information:
\begin{equation}
\label{eq:graphconv}
\begin{split}
   {\hat{\mathbf{H}}^{m}}  =  [ \hat{\mathbf{h}}_{1}^{m}, \hat{\mathbf{h}}_{2}^{m} & , \,\dots ,  \hat{\mathbf{h}}_{B}^{m}]^{\intercal}  = \operatorname{GCN}({\mathbf{H}}^{m}, \tilde{\Pi}) \\
   & = \operatorname{ReLU}(\tilde{\Pi} \  \operatorname{ReLU}(\tilde{\Pi} {\mathbf{H}}^{m} W^0)W^1 ) ,
\end{split}
\end{equation}
where ${\hat{\mathbf{H}}^{m}}$ is the aggregated auxiliary information from similar patients in the space of modality $m$.
$W^0$ and $W^1\in\mathbb{R}^{N_{h} \times N_{h}}$ are the projection matrices.
We ignore the bias term here and after.

\subsection{Adaptive Modality Imputation}

Now we obtain two different representations for the batch of patients in each modality: $\mathbf{H}^{m}$ and ${\hat{\mathbf{H}}^{m}}$. 
The former focuses on the patients themselves for modality $m$, while the latter refers to the information aggregated from similar patients.
For a specific patient $i$, if the data of modality $m$ are missing, we can directly impute the representation with the auxiliary information aggregated from similar patients.
While for a patient whose modality $m$ is complete, such auxiliary information can also be fused into the original representation, making the representation smoother to reduce noise, thus enhancing the representation learning.
Here, we use an attention fusion to adaptively extract the proper amount of information from them (i.e., ${\mathbf{H}}^{m}$ and ${\hat{{\mathbf{H}}}^{m}}$).
Specifically, two weights $\alpha^{m}, \beta^{m} \in\mathbb{R}^{B \times 1}$ are introduced to determine the importance of the above two representations, which are obtained by fully connected layers:
\begin{equation}
\label{eq:alpha}
    \alpha^{m}=\operatorname{Sigmoid}(  {\mathbf{H}}^{m} W_{o}), \beta^{m}=\operatorname{Sigmoid}({ \hat{{\mathbf{H}}}^{m}} W_{s} ),
\end{equation}
where $ W_{o}, W_{s} \in\mathbb{R}^{ N_{h}  \times 1}$ are the weight matrices.
$\alpha$ and $\beta$ indicate the importances of self-information and information of similar patients. 
We add a constraint $\alpha + \beta = 1$ by calculating $ \alpha = \frac{\alpha}{\alpha + \beta},$ $\beta = 1 - \alpha.$
The final imputed and enhanced representations can be obtained as (i.e., the No.3 arrow in Figure~\ref{fig:framework}):
\begin{equation}
\label{eq:impute}
\mathbf{h}_{i}^{m} = 
\begin{cases}
 \qquad\quad\ \  \hat{\mathbf{h}}_{i}^{m}&  \text{if modality } m \text{ of sample }  i \text{ is missing}  \\
 \alpha^{m}_{i} \cdot \mathbf{h}_{i}^{m} + \beta^{m}_{i} \cdot  \hat{\mathbf{h}}_{i}^{m} & \text{otherwise}

\end{cases}
\end{equation}
where $\mathbf{h}_{i}^{m} $ and  $\hat{\mathbf{h}}_{i}^{m} $ are the $i$-th sample of ${\mathbf{H}}^{m}$ and ${ \hat{{\mathbf{H}}}^{m}}$, respectively.

\subsection{Multimodal Interaction Capture}
Back to a specific patient, so far, the representations of the missing modalities have been imputed
through the above sections.
These representations are used to perform the clinical tasks.
Thus, we need to consider complex correlations among multimodal EHR, including intra-correlations within each modality and inter-correlations between modalities.
Inspired by~\citet{kim2021vilt,akbari2021vatt}, a context-aware multimodal interaction capture is built.
Specifically, for sequential modalities, the internal positional encoding is added: $\bar{\mathbf{H}}^{\text{seq}} = \left[ \mathbf{h}^{\text{seq}}_{1}, \mathbf{h}^{\text{seq}}_{2}, \,\dots , \mathbf{h}^{\text{seq}}_{N_{\text{seq}}}  \right]  + \text{PE}^{\text{seq}},$
where $\text{PE}^{\text{seq}}$ is the positional encoding for sequential modality $\text{seq}$ and $N_{\text{seq}}$ is the length of the sequence.
Next, the representations are added with the corresponding modality type embeddings and concatenated to form the input:
\begin{equation}
\label{eq:mmbegin}
\mathbf{z}^{0} = [\bar{\mathbf{H}}^{1}+\text{TE}^{1}; \bar{\mathbf{H}}^{2}+\text{TE}^{2} ;... ; \bar{\mathbf{H}}^{M}+\text{TE}^{M}],
\end{equation}
where $\text{TE}^{m}$ is the corresponding type embedding to identify each modality.
And the multimodal interactions are captured through:
\begin{equation}
\label{eq:mmend}
\begin{split}
   \tilde{\mathbf{z}}^{l} & = \operatorname{LayerNorm}({\mathbf{z}}^{l-1} + \operatorname{MHSA}({\mathbf{z}}^{l-1})), \\
   \mathbf{z}^{l}  &= \operatorname{LayerNorm}(\tilde{\mathbf{z}}^{l} + \operatorname{FFN}(\tilde{\mathbf{z}}^{l})), \\
\end{split}
\end{equation}
where $l = 1, \ ..., L $ refers to the number of such stacked layers.
MHSA refers to the Multi-Head Self-Attention~\cite{vaswani2017attention}, FFN refers to a feed-forward network and LayerNorm is the layer normalization ~\cite{ba2016layer}.
The predictor is built via:
\begin{equation}
\label{eq:predict1}
    \hat{\mathbf{y}}_i = \operatorname{Sigmoid}( \mathbf{z}^{L}_{0}  \, W_{final}),
\end{equation}
where $ W_{final} \in\mathbb{R}^{ N_{h} \times 1}$ is the weight matrix.
The Sigmoid function is used to turn the output into the probability.
In this case, the cross-entropy loss is used as the prediction loss function:
\begin{equation}
\mathcal{L}_{\text{pre}} = -\frac{1}{B} \sum_{i=1}^{B}(\mathbf{y}_i^{\intercal} \operatorname{log}(\hat{\mathbf{y}}_i) + (1-\mathbf{y}_i )^{\intercal} \operatorname{log}(1-\hat{\mathbf{y}}_i)),
\end{equation}
where $B$ is the batch size. 
$\hat{\mathbf{y}}_i \in [0, 1]^{|C|} $ is the predicted probability, and ${\mathbf{y}}_i \in \{0, 1\}^{|C|} $ is the ground truth.
The overall loss function is:
\begin{equation}
\label{eq:loss}
\mathcal{L} = \mathcal{L}_{\text{pre}} + \lambda \mathcal{L}_{\text{stab}},
\end{equation}
where $\lambda$ is the hyperparameter to control the loss.
For ease of understanding, we summarize \mname in Algorithm~\ref{alg} in Appendix.

\section{Experiment}

We evaluate \mname on the following datasets: Ocular Disease Intelligent Recognition (ODIR) Dataset and Ophthalmic Vitrectomy (OV) Dataset. The model code is provided in~\footnote{\url{https://github.com/choczhang/M3Care}}.

\subsection{Data Description and Task Formulation}

We use the following datasets and tasks to evaluate our model.

\begin{itemize}[leftmargin=*,noitemsep,topsep=3pt]

\item
\textbf{Ocular Disease Intelligent Recognition (ODIR) Dataset} comes from an ophthalmic database, which is meant to represent real-life set of patients collected from hospitals~\cite{li2021benchmark}.
3,500 patients are extracted to construct this dataset to diagnose ocular diseases.
This dataset contains the following modalities (incomplete modalities exist): demographic information, clinical text for both eyes, and fundus images for both eyes.
The detailed statistics are presented in Table~\ref{tab:dataset} in Appendix.

\textbf{Task.} The ocular diseases diagnosis task on this dataset is defined as a multi-label classification task.
Following existing works~\cite{bai2019improving,yu2020experimental}, we assess the performance using micro-averaged of the area under the receiver operating characteristic curve (i.e., micro-AUC), macro-AUC, and the average test loss value.
We divide the dataset into the training set, validation set, and test set with a proportion of 0.8\,:\,0.1:\,0.1, and report the performance with the standard deviation of bootstrapping for 1,000 times.
\item
\textbf{Ophthalmic Vitrectomy (OV) Dataset} comes from an ophthalmic hospital\footnote{This study was approved by the Research Ethical Committee.}.
In clinical practice, after vitrectomy, the intraocular pressure (IOP) may increase abnormally. 
This symptom cannot be predicted by doctors. 
We collect 832 patients to predict whether the IOP will increase abnormally. 
This dataset contains six modalities (incomplete modalities exist): demographic,  clinical notes, medications, admission records, discharge records, and surgical consumables.
The detailed statistics are presented in Table~\ref{tab:dataset}.

\textbf{Task.} The task on the dataset are 
binary classification tasks.
Following existing works~\cite{zhang2021grasp,hoang2021aid}, we assess performance using the area under the precision-recall curve (AUPRC), the area under the ROC curve (AUROC), and accuracy (ACC).
AUPRC is the most informative and primary evaluation metric, especially while dealing with skewed real-world data~\cite{davis2006relationship,choi2018mime,zhang2021grasp}.
Due to the size of the dataset, we employ 10-fold cross-validation to assure the consistency of the performance and report the average performance with standard deviations.

\end{itemize}

\subsection{Experimental Setup and Baselines}

To conduct the experiment, we use the Adam optimization algorithm in Pytorch 1.5.1. 
More \hyperref[sec:detailed]{details are in the Appendix}.
We include these state-of-the-art models as our baseline models:

\begin{itemize}[leftmargin=*,noitemsep,topsep=1pt]

\item \textbf{MFN}~\cite{zadeh2018memory} captures view-specific and cross-view interactions, and summarizes them with a multi-view gated memory module.

\item \textbf{MulT}~\cite{tsai2019multimodal} utilizes directional pairwise cross-modal transformers to attend to interactions between multimodal data.

\item \textbf{ViLT}~\cite{kim2021vilt} commissions the transformer module to extract and process all the multimodal features simultaneously.


\item \textbf{CM-AEs}~\cite{ngiam2011multimodal}: The cross-modal autoencoders, which generate missing modalities first and make predictions.


\item \textbf{SMIL}~\cite{ma2021smil} approximates the missing modality using a weighted sum of manually defined modality priors learned from the dataset. 

\item \textbf{HGMF}~\cite{chen2020hgmf} fuses incomplete multimodal data within a heterogeneous graph structure, and we modify it to an inductive version. 



\end{itemize}

The following ablation studies are also conducted:
\begin{itemize}[leftmargin=*,noitemsep,topsep=1pt]
    \item \textbf{$\mname_{1-}$} does not use the task-guided deep kernels of each modality. It directly calculates similarity via cosine similarity.
    \item \textbf{$\mname_{2-}$}  does not consist of the Information Aggregation and the Adaptive Modality Imputation module. 
    It directly computes the mean similarity from each modality and approximates the missing-modality representations via the similar patients.
\end{itemize}

It should be noted that some of the above models' embedding networks of raw data are a little bit weak. 
Thus, to perform a fair comparison, we upgrade their embedding layers to the same ones as ours (e.g., Transformer Encoder~\cite{vaswani2017attention} and ResNet18~\cite{he2016deep}) and we do not include any pre-trained parameters.

\subsection{Experimental Results}

As shown in Table~\ref{tab:result_ODIR} and \ref{tab:result_OV}, we can see that \mname can outperform all the baselines in terms of different evaluation metrics\footnote{${}^{**}:p<0.01$, ${}^{*}:p<0.05$}.

\begin{table}[h]
\small 
  \centering
  \caption{Results on the ODIR Dataset}
  \label{tab:result_ODIR}

\begin{tabular}{cccc}
\hline

 & \multicolumn{3}{c}{\textbf{ODIR Dataset} (Multi-label Classification) }\\

 \textbf{Methods}&\textbf{micro-AUC$\ \uparrow$}&\textbf{macro-AUC$\ \uparrow$}&\textbf{test loss$\ \downarrow$}\\ 
\hline

MFN~\cite{zadeh2018memory}   & 0.7877\,(0.030) & 0.7766\,(0.029) & 0.1772\,(0.020) \\ 
MulT~\cite{tsai2019multimodal}  & 0.7944\,(0.028) & 0.8032\,(0.026) & 0.2339\,(0.019)  \\ 
ViLT~\cite{kim2021vilt}   & 0.7966\,(0.031) & 0.7624\,(0.029) & 0.1731\,(0.016) \\
CM-AEs~\cite{ngiam2011multimodal} & 0.8028\,(0.030) & 0.7672\,(0.027) & 0.1878\,(0.031) \\ 
SMIL~\cite{ma2021smil}   & 0.8092\,(0.032) & 0.7978\,(0.025) & 0.2278\,(0.032) \\ 
HGMF~\cite{chen2020hgmf} & 0.8080\,(0.030) & 0.8103\,(0.031) & 0.1810\,(0.022) \\
\hline
 $\mname_{1-}$ & 0.8130\,(0.031) & 0.8059\,(0.032) & 0.1781\,(0.020)  \\ 
   $\mname_{2-}$  & 0.8030\,(0.031) & 0.8138\,(0.029) & 0.1631\,(0.018)  \\ 
\hline
 \textbf{ $\mname$} & \textbf{0.8490}$^{**}$\,(0.025) & \textbf{0.8245}$^{**}$\,(0.026) & \textbf{0.1543}$^{**}$\,(0.018)   \\ 
\hline
\end{tabular}
\end{table}

\begin{table}[h]
\small
  \centering
  \caption{Results on the OV Dataset}
  \label{tab:result_OV}

\begin{tabular}{cccc}
\hline

 &  \multicolumn{3}{c}{\textbf{OV Dataset} (Binary Classification) }\\

 \textbf{Methods}& \textbf{AUPRC$\ \uparrow$}& \textbf{AUROC$\ \uparrow$}& \textbf{ACC$\ \uparrow$}  \\ 
\hline

MFN~\cite{zadeh2018memory}   & 0.6456\,(0.038) & 0.6789\,(0.032)  & 0.6627\,(0.032) \\ 
MulT~\cite{tsai2019multimodal}  & 0.6814\,(0.047)& 0.6891\,(0.043)  & 0.6988\,(0.031) \\ 
ViLT~\cite{kim2021vilt}  & 0.6987\,(0.051) & 0.7245\,(0.048)  & 0.6627\,(0.033) \\ 
CM-AEs~\cite{ngiam2011multimodal} & 0.6891\,(0.031) & 0.6927\,(0.040) & 0.6747\,(0.029) \\ 
SMIL~\cite{ma2021smil}    & 0.7109\,(0.045)& 0.7041\,(0.033)  & 0.6867\,(0.032) \\ 

HGMF~\cite{chen2020hgmf} & 0.7037\,(0.050) & 0.7544\,(0.027) & 0.7100\,(0.032) \\
\hline
 $\mname_{1-}$ & 0.6849\,(0.054) & 0.7472\,(0.052)  & 0.7080\,(0.064) \\ 
   $\mname_{2-}$ & 0.7110\,(0.044) & 0.7562\,(0.057)  & 0.7234\,(0.072) \\ 
\hline
     
 \textbf{ $\mname$}  & \textbf{0.7549}$^{**}$\,(0.065) & \textbf{0.7998}$^{**}$\,(0.049)  & \textbf{0.7438}$^{**}$\,(0.058) \\ 
\hline
\end{tabular}
\end{table}

Specifically, on the ODIR Dataset, the number in () denotes the standard deviation of bootstrapping for 1,000 times.
The results show that, compared with the best baseline method, \mname achieves relative improvements of 4.9\% in micro-AUC.
On the OV Dataset, the number in () denotes the standard deviation of 10-fold cross-validation.
We can see that, compared with the best baseline method, \mname achieves relative improvements of 6.1\% in AUPRC and 6.0\% in AUROC.
Among these baseline methods, some ones like CM-AEs~\cite{ngiam2011multimodal},  SMIL~\cite{ma2021smil}, HGMF~\cite{chen2020hgmf} use various mechanisms to handle the missing modalities, and thus they achieve relative higher performance.
However, the performance boost demonstrates the effectiveness of \mname.
Besides, it is worth mentioning that the OV Dataset only contains the multimodal EHR data of 832 patients yet has six modalities, demonstrating that \mname performs well on a small dataset while the number of modalities is large, which is suitable for the real-world scenario, where data has a large number of modalities or missing patterns.

The superior performance of \mname than the $\mname_{1-}$ (i.e., calculating similarity via cosine similarity) verifies the efficacy of the task-guided deep kernels. 
Moreover, \mname outperforms $\mname_{2-}$, which demonstrates the superiority of the Information Aggregation and the Adaptive Modality Imputation module.

\subsection{Further Analysis}

We conduct several further experiments.
Due to the limitation of pages, some of the experiments are in the Appendix.

\subsubsection{Clinical implications} \label{sec:cli_imp}


\begin{figure}
 \centering 
 \subfigure[]{\includegraphics[width=1.6in]{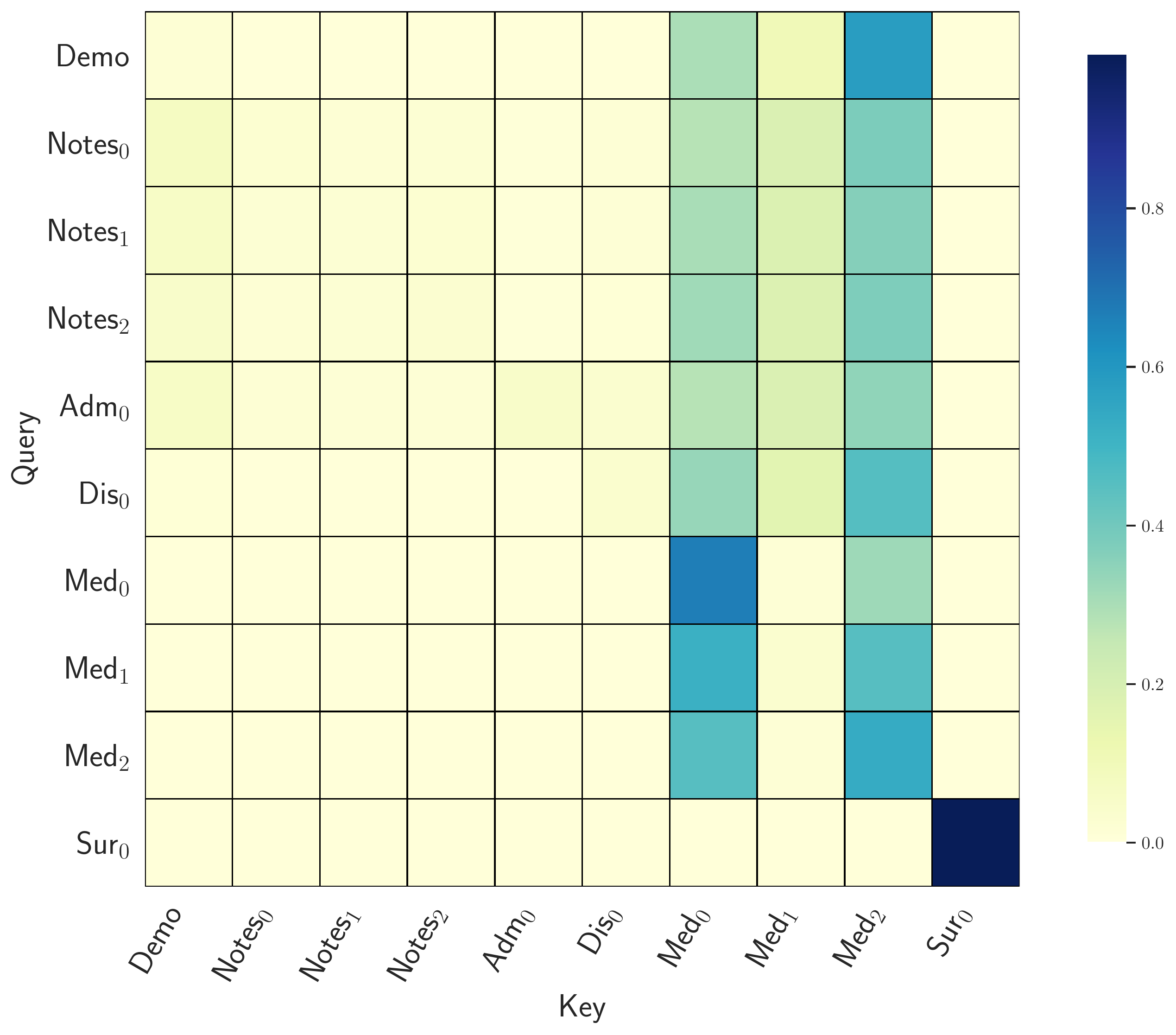}}
 \subfigure[]{\includegraphics[width=1.6in]{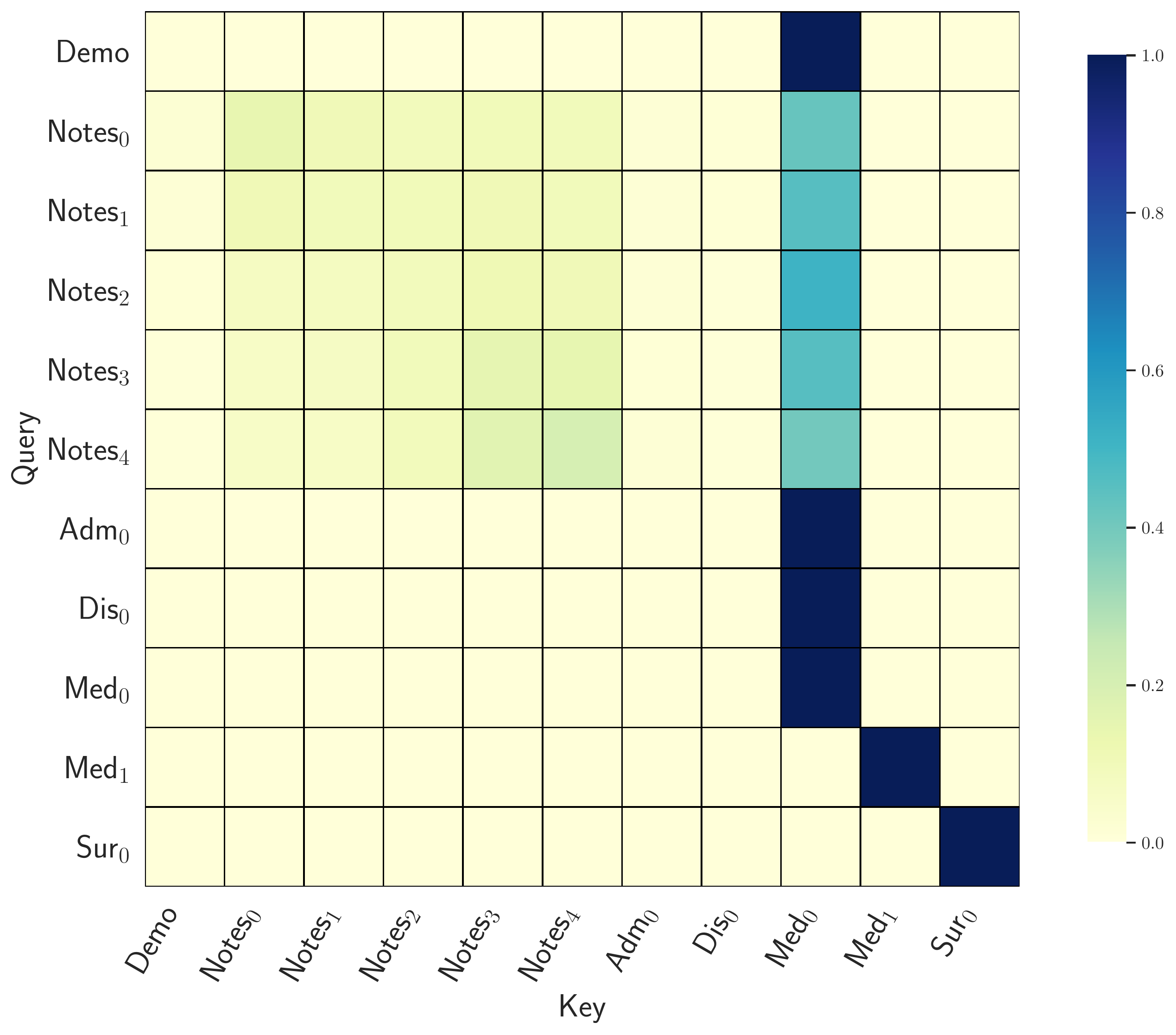}}
 \caption{Attention weight visualization for two patients with abnormally increased intraocular pressure in the test set of the OV dataset.~\textit{Best viewed in color.}}
 \label{fig:case}
\end{figure}

To intuitively show the implication of \mname, we visualize the attention weights of the prediction process.
Due to the limitation of pages, we report two cases in the test set here.
As shown in Figure~\ref{fig:case}, the intraocular pressure (IOP) of the two patients increase abnormally and \mname successfully predicts the outcome.
The rows and columns show the Query and Key multimodal records, which are the abbreviations of each modality, i.e., demographic information, clinical notes, medications, admission records, discharge records, and surgical consumables, respectively.

We notice that \mname gives strong focus on $\text{Med}_0$ and $\text{Med}_2$  (i.e., the first and third medications) of patient $a$, and $\text{Med}_0$ of patient $b$.
In all these medications, the patients received the two drugs: \textit{Tropicamide Phenylephrine Eye Drops (Mydrin-P)} and \textit{Prednisolone Acetate Ophthalmic Suspension (Pred Forte)}.
These drugs are used for ophthalmologic examinations, prior to ocular surgery~\cite{Mydrin} or treat eye swelling caused by allergy, infection, injury, or other conditions~\cite{PRED_FORTE}.
Our model discovers that these two drugs may have a strong relationship with the abnormally increased intraocular pressure.
This is highly consistent with medical literature~\cite{kim2012changes,atalay2015change,matossian2020impact,rajendrababu2021incidence} and clinician experience, which confirm that the two drugs can lead to adverse reactions like elevation of IOP and should be used with caution for specific patients in clinical practice.



\section{conclusions}
In this paper, we propose \mname, an end-to-end model to compensate for the missing information of the patients with missing modalities and perform clinical prediction as well as analysis.
For a patient with missing modalities, \mname finds similar patients with a task-guided modality-adaptive similarity metric. 
Instead of generating raw missing data, \mname imputes the hidden representations of the missing modalities in the latent space by the auxiliary information from these similar ones, and conducts the clinical tasks.
Experiments show that \mname outperforms all baseline models. 
Besides, the findings are in accord with experts and medical knowledge, which shows it can provide useful insights.

\begin{acks}
This work is supported by the National Natural Science Foundation of China (No.62172011).
L. Ma is supported by the China Postdoctoral Science Foundation (2021TQ0011).
J. Wang is supported by EPSRC New Investigator Award under Grant No.EP/V043544/1.
\end{acks}

\bibliographystyle{ACM-Reference-Format}
\balance
\bibliography{sample-sigconf}

\clearpage

\appendix

\section{Intuition discovery experiment}
\label{sec:discovery}
In each dataset, for each modality, we first take the samples containing the complete modality data and divide them into training, validation, and test sets. 
Then we use a unimodal classifier (e.g., multilayer perceptron, transformer encoder) to classify the data for each modality. 
We take the best model from the validation set, apply it to the test set, and collect the representations in the latent space of each modality for each sample.
Next, we compute pair-wise similarity matrix $\Pi_m$ between samples in each modality, where $m$ is the corresponding modality.

We want to justify the intuition in our datasets: if two patients are similar in one modality, they are more likely to be similar in another modality with regard to the clinical task.
We call intuition the \textit{cross-modal transfer of sample similarity}.
Given modalities $a$ and $b$, the intuition holds if their difference of the pair-wise similarity matrix $\|\Pi_a - \Pi_b \|_\text{norm}$ is small enough, where $\text{norm}$ is a type of matrix norm.
In this experiment, we try different similarity metrics such as normalized Euclidean distance, cosine similarity and RBF kernel.
And we also try different norms as metrics of difference, such as Frobenius norm, 2-norm and mean value of all entries in $\|\Pi_a - \Pi_b \|$.

For comparison, we add noise to the representations in the latent space in one modality (we select modality $b$ here), and calculate the difference $\|\Pi_a - \Pi_b^{'} \|_\text{norm}$.
If the intuition holds, the difference should be bigger than the above original $\|\Pi_a - \Pi_b \|_\text{norm}$.
We perform this experiment 1,000 times and calculate the average difference to avoid chance.

Furthermore, we also shuffle the representations in the latent space in one modality (we select modality $b$ here), and calculate the difference $\|\Pi_a - \Pi_b^{''} \|_\text{norm}$.
If the intuition holds, the difference should also be bigger than the first original $\|\Pi_a - \Pi_b \|_\text{norm}$.
In the same way, we perform this experiment 1,000 times and calculate the average difference to avoid chance.
The results are shown in Table
~\ref{tab:ov_intuition1} and~\ref{tab:ov_intuition2}.

\begin{table}[b]
  \centering
  \caption{Intuition observation results for two modalities: admission records and clinical notes, on the OV Dataset}
  \label{tab:ov_intuition1}
\begin{tabular}{ccccc}
\hline

Metric                                       & Norm      & Original & Noise   & Shuffle \\ \hline
\multicolumn{1}{c}{\multirow{3}{*}{\begin{tabular}[c]{@{}c@{}}Normalized \\ Euclidean \\ Distance\end{tabular}}} & Frobenius & 183.35   & 403.23 & 207.27  \\
\multicolumn{1}{c}{}                         & 2         & 153.97   & 385.21 & 173.73  \\
\multicolumn{1}{c}{}                         & Mean      & 0.2334   & 0.4871 & 0.2724  \\ \hline
\multirow{3}{*}{\begin{tabular}[c]{@{}c@{}}Cosine \\ similarity\end{tabular}} & Frobenius & 402.96   & 462.59 & 452.72  \\
                                             & 2         & 336.31   & 365.02 & 376.69  \\
                                             & Mean      & 0.5081   & 0.5821 & 0.5908  \\ \hline
\multirow{3}{*}{\begin{tabular}[c]{@{}c@{}}RBF \\ kernel\end{tabular}}   & Frobenius & 176.35   & 261.05 & 199.60  \\
                                             & 2         & 147.51   & 196.10 & 166.46  \\
                                             & Mean      & 0.2241   & 0.3024 & 0.2623  \\ \hline
\end{tabular}
\end{table}

As shown in Table~\ref{tab:ov_intuition1}, the original difference of the pair-wise similarity matrices in the two modalities is smaller than both the Noise and Shuffle ones with regard to different similarity metrics and different norms.
This justifies that if two patients are similar in one modality, they are more likely to be similar in another modality in different view.
To this end, we come up with our intuition, i.e., the \textit{cross-modal transfer of sample similarity}.
In another pair, as shown in Table~\ref{tab:ov_intuition2}, the same conclusion can be drew.

\begin{table}[b]
  \centering
  \caption{Intuition observation results for two modalities: medications and surgical consumables information, on the OV Dataset}
  \label{tab:ov_intuition2}
\begin{tabular}{ccccc}
\hline

Metric                                       & Norm      & Original & Noise   & Shuffle \\ \hline
\multicolumn{1}{c}{\multirow{3}{*}{\begin{tabular}[c]{@{}c@{}}Normalized \\ Euclidean \\ Distance\end{tabular}}}  & Frobenius & 152.96   & 159.02 & 370.39  \\
\multicolumn{1}{c}{}                         & 2         & 131.37   & 145.77 & 360.48  \\
\multicolumn{1}{c}{}                         & Mean      & 0.2021   & 0.2289 & 0.5588  \\ \hline
\multirow{3}{*}{\begin{tabular}[c]{@{}c@{}}Cosine \\ similarity\end{tabular}}& Frobenius & 384.57   & 406.01 & 404.89  \\
                                             & 2         & 326.61   & 353.72 & 367.75  \\
                                             & Mean      & 0.4984   & 0.5328 & 0.5671  \\ \hline
\multirow{3}{*}{\begin{tabular}[c]{@{}c@{}}RBF \\ kernel\end{tabular}} & Frobenius & 168.85   & 173.56 & 203.48  \\
                                             & 2         & 144.43   & 157.96 & 188.48  \\
                                             & Mean      & 0.2081   & 0.2260 & 0.2825  \\ \hline
\end{tabular}
\end{table}

\section{algorithm}

Algorithm~\ref{alg} shows the algorithm of \mname.

\begin{algorithm}[]
\SetAlgoLined
\KwInput{\\Multimodal EHR dataset $\mathbf{\mathbb{X}}$ }
\KwOutput{\\Prediction for the patient $\hat{\mathbf{y}}$}
\KwTraining{}
 Initialize weights\;
 \While{training is not convergence}{
  \For {each batch of patient }{
 Extract $\mathbf{h}^{m}_{n}$ of each modality $m$ via Eq.~\ref{eq:hm}\;
 Form the batch-wise representation matrices $\mathbf{H}^{m}$\;
 Compute patient similarity matrix $\Pi^{m }$ in each modality space via Eq.~\ref{eq:pim}\;
 Compute  comprehensive similarity $\tilde{\Pi} $ via Eq.~\ref{eq:pi},~\ref{eq:piij}\;
 Form the similar patient graph with ${\mathbf{H}}^{m}$ as nodes and $\tilde{\Pi} $ as adjacency matrix\;
 Compute the aggregated information ${\hat{\mathbf{H}}^{m}}$  via Eq.~\ref{eq:graphconv} in the space of each modality $m$\;
 \For {each patient in the batch}{
 \If {missing modalities exist}{Impute the missing modalities via Eq.~\ref{eq:impute}}
 \Else{Enhance the representations via Eq.~\ref{eq:alpha}-\ref{eq:impute}\;}

 Model the multimodal dynamics via Eq.~\ref{eq:mmbegin}-\ref{eq:mmend}\;
 Make prediction via Eq.~\ref{eq:predict1}\;

 }
 Update the parameters by optimizing Eq.~\ref{eq:loss}\;
 }
 }
 \caption{Algorithm of \mname}\label{alg}
\end{algorithm}

\section{Further Analysis}
\subsection{Multiple levels multimodal incompleteness}

\begin{table}[]
\small 
  \centering
  \caption{Micro-AUC on ODIR Dataset under different additional synthesizing multimodal missing rates.}
  \label{tab:result_missing_odir}
\begin{tabular}{c|cccc}
\hline


 \textbf{Methods}&30\%&40\%&50\%&60\%\\ 
\hline

MFN~\cite{zadeh2018memory}  & .7625\,(.03) & .7334\,(.03)  & .7260\,(.03)  & .7166\,(.03) \\ 
MulT~\cite{tsai2019multimodal} & .7768\,(.02) & .7637\,(.03)  & .7480\,(.03)  & .7348\,(.03) \\ 
ViLT~\cite{kim2021vilt}  & .7601\,(.03) & .7583\,(.03)  & .7492\,(.03)  & .7355\,(.02) \\  
CM-AEs~\cite{ngiam2011multimodal} & .7846\,(.03) & .7707\,(.00)  & .7648\,(.03)  & .7477\,(.03) \\ 
SMIL~\cite{ma2021smil}   & .7702\,(.02) & .7595\,(.02)  & .7485\,(.03)  & .7396\,(.03) \\  
HGMF~\cite{chen2020hgmf}  & .7831\,(.03) & .7711\,(.03)  & .7585\,(.03)  & .7427\,(.02) \\ 

\hline
 \textbf{ $\mname$} & \textbf{.8119}$^{**}$\,(.02) & \textbf{.7927}$^{**}$\,(.03) & \textbf{.7795}$^{**}$\,(.03) & \textbf{.7715}$^{**}$\,(.02)  \\ 
\hline
\end{tabular}
\end{table}

To consider more realistic various settings and verify the generalizability of \mname, experiments on multimodal data with multiple levels of multimodal incompleteness are conducted.
We evaluate the influences of missing modalities by attaching additional synthesizing multimodal incompleteness rates on ODIR Dataset from 30\% to 60\% with an intermittent 10\%. 
The experiments are repeated test with bootstrapping for 1,000 times, and the results (micro-AUC) are in Table~\ref{tab:result_missing_odir}.


We can see that all the models' micro-AUCs decrease as the missing rate increases, and \mname still outperforms all baselines.
When the missing rate is the biggest of all settings (60\%), \mname also demonstrated significantly better performance than the best baselines CM-AEs~\cite{ngiam2011multimodal} and HGMF~\cite{chen2020hgmf}.
Specifically, \mname achieves a micro-AUC of 0.7715, while the baseline models CM-AEs~\cite{ngiam2011multimodal} and HGMF~\cite{chen2020hgmf} achieve 0.7477 and 0.7427, showing 3.2\% and 3.8\% relative improvement, respectively.

\subsection{Clinical implications}


\begin{figure}[]
 \centering
 \subfigure[a positive case]{\includegraphics[width=1.6in]{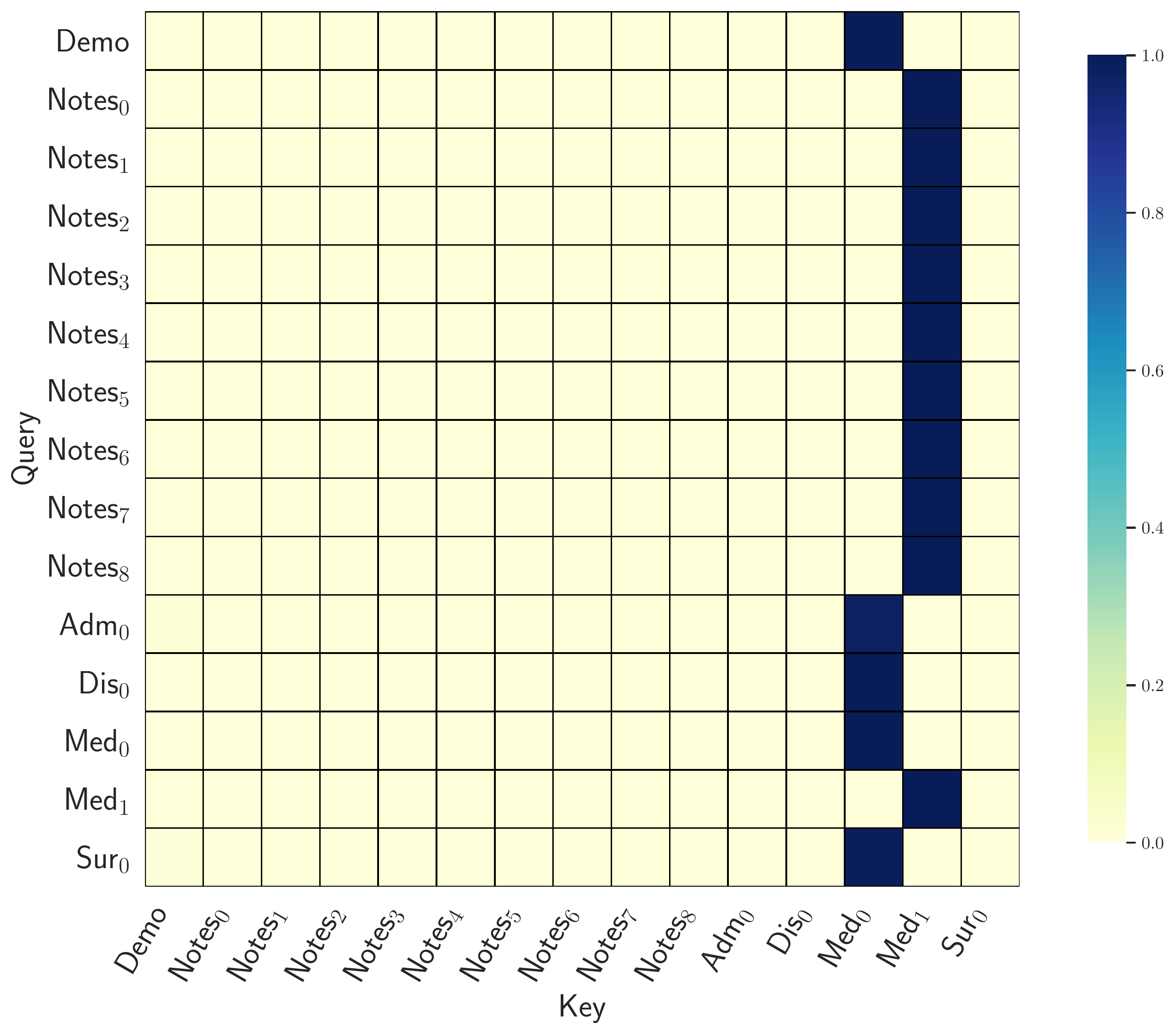}}
 \subfigure[a negative case]{\includegraphics[width=1.6in]{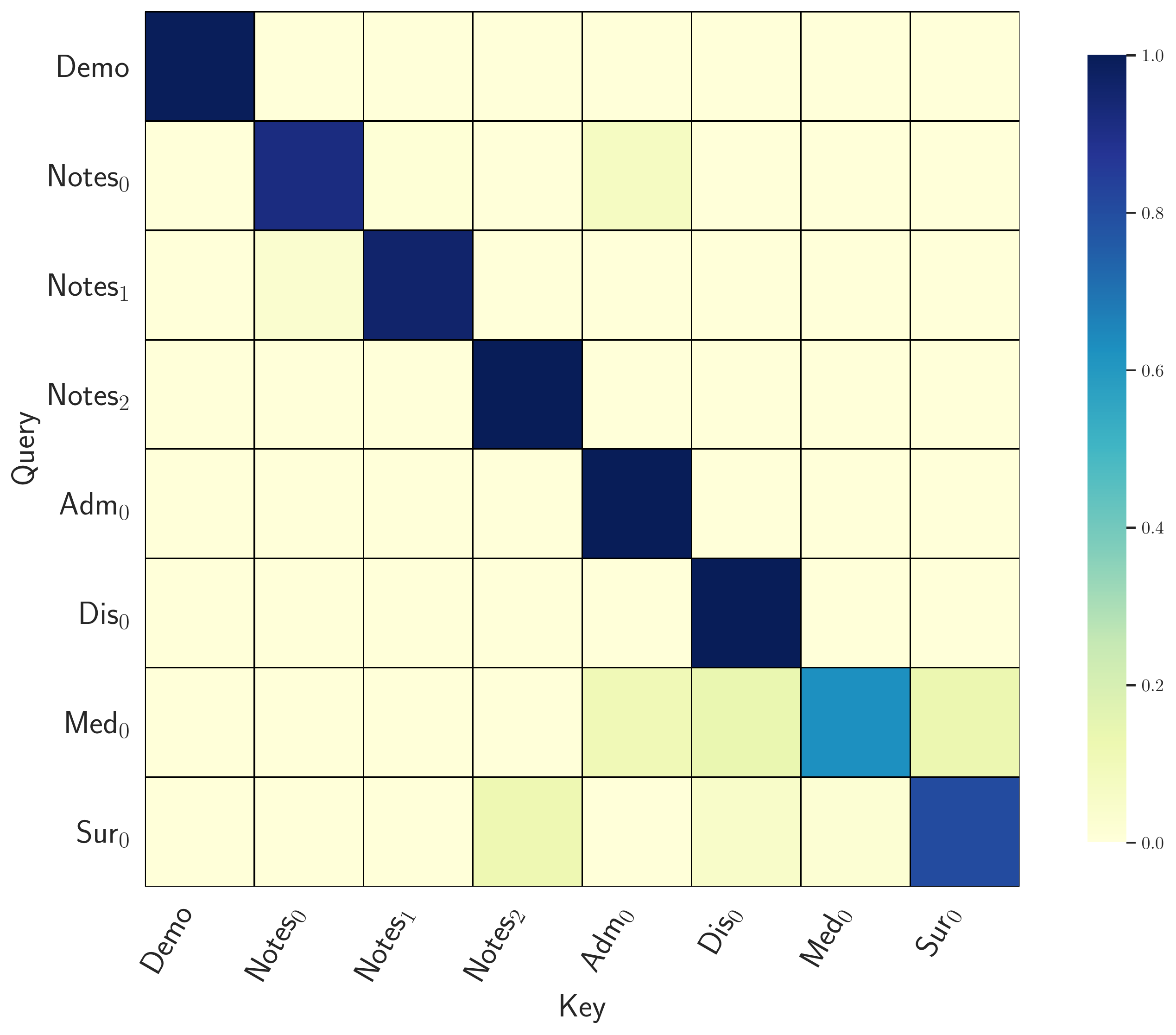}}
 \caption{Attention weight visualization for two patients in the test set of OV dataset.}
 
 \label{fig:case2}
\end{figure}

To intuitively show the implication of \mname, similar to Section~\ref{sec:cli_imp}, we further visualize the attention weights of the prediction process for another two cases.
As shown in Figure~\ref{fig:case2}, the first case has a positive label, and the second one is negative.
For the first one, \mname gives a strong focus on $\text{Med}_0$ and $\text{Med}_1$.
In both these medications, the patients received not only the drug mentioned above in Section~\ref{sec:cli_imp}: \textit{Tropicamide Phenylephrine Eye Drops (Mydrin-P)}, which have been proved by medical literature~\cite{kim2012changes,atalay2015change} and clinician experience that they can lead to adverse reactions like elevation of IOP.
The patients also received Tobramycin Dexamethasone Eye Drops (Tobradex), which is highly consistent with medical literature~\cite{chen2016comparison}
and clinician.
The drugs can lead to adverse reactions and have the tendency to increase intraocular pressure~\cite{chen2016comparison}.
For the second one, a healthy patient with a negative label, \mname gives relatively even attention to each data of each modality of the patient, which indicates that \mname does not discover significant signs of elevation of IOP and finally makes a right prediction.




\section{Details of Experimental Settings}
\label{sec:detailed}

\subsection{Statistics of the Datasets}

\begin{itemize}[leftmargin=*,noitemsep,topsep=3pt]

\item
\textbf{Ocular Disease Intelligent Recognition (ODIR) Dataset} contains the following modalities (incomplete modalities exist): demographic information, clinical text for both eyes, and fundus images for both eyes.
The detailed statistics are presented in Table~\ref{tab:dataset}.

\item
\textbf{Ophthalmic Vitrectomy (OV) Dataset} contains six modalities (incomplete modalities exist): demographic,  clinical notes, medications, admission records, discharge records, and surgical consumables.
The detailed statistics are presented in Table~\ref{tab:dataset}.

\end{itemize}

\begin{table}[h]
\small
\caption{Statistics of the Datasets}
\label{tab:dataset}
\begin{tabular}{llc}
\hline
Dataset                                                                                                                 & Statistic                    & Value                     \\ \hline

\multirow{5}{*}{\begin{tabular}[c]{@{}l@{}}Ocular Disease \\ Intelligent \\ Recognition \\ (ODIR) Dataset\end{tabular}} & \# patients                   & 3,500                      \\
                                                                                                                        & \# modalities                & 3                         \\
                                                                                                                      & \% missing per modality
                                                                                                                      & {[}0\%, 48.34\%, 0\%{]}   \\
                                                                                                                        & \% positive labels           & {[}0.061, 0.060, .0046{]} \\
                                                                                                                        
                                                                                                                        & \% female                    & .461                     \\ \hline
\multirow{6}{*}{\begin{tabular}[c]{@{}l@{}}Ophthalmic \\ Vitrectomy \\ (OV) Dataset\end{tabular}}                       & \# patients                   & 832                       \\
                                                                                                                        & \#  modalities               & 6                         \\
                                                                                                                        & \% missing per modality\tablefootnote{The order of the missing rates are
                                                                                                                      corresponding to the above data description.}      &  {[}0\%, 0.60\%, 15.38\%, \\
                                                                                                                        &  &10.33\%, 10.33\%, 4.08\%{]} \\
                                                                                                                        & \% positive labels           & 41.7\%                      \\
                                                                                                                        
                                                                                                                        & \% female                    & .456                     \\ \hline

                                                                                                                        

\end{tabular}
\end{table}

\subsection{Model Implementation}

The experiment environment is a machine equipped with CPU: Intel Xeon E5-2630, 256GB RAM, and GPU: Nvidia RTX8000. The code is implemented based on Pytorch 1.5.1. 
The hyper-parameter setting of the proposed \mname is as follows:
We set the embedding dimension and hidden dimension as 128/256 for Ocular Disease Intelligent Recognition (ODIR) / Ophthalmic Vitrectomy (OV) dataset, respectively.
Since the clinical notes modality in OV dataset is too long per patient, and the number of samples is small (832 patients), we set the batch size as 32 while conducting experiments on OV dataset.
For ODIR dataset, we set the batch size as 512.

\end{document}